%% file: main.tex
\documentclass[10pt,twocolumn,letterpaper]{article}

 \usepackage{iccv}              


\definecolor{iccvblue}{rgb}{0.21,0.49,0.74}
\usepackage[pagebackref,breaklinks,colorlinks,allcolors=iccvblue]{hyperref}

\def\paperID{1} 
\def\confName{ICCV}
\def\confYear{2025}

\title{AI vs. Human Moderators: A Comparative Evaluation of Multimodal LLMs in Content Moderation for Brand Safety \thanks{Accepted to the Computer Vision in Advertising and Marketing (CVAM) workshop at ICCV 2025.}}

\author{
Adi Levi \quad Or Levi \quad Sardhendu Mishra \quad Jonathan Morra \\
Zefr Inc \\
{\tt\small adi.levi@zefr.com, or.levi@zefr.com, sardhendu.mishra@zefr.com, jon.morra@zefr.com}
}

\newcommand{\modelname}{MLLMs}  
\newcommand{\modelnamefull}{Multimodal Large Language Models}  

\begin{document}
\maketitle
\input{sec/0_abstract}    
\input{sec/1_intro}
\input{sec/2_related_work}
\input{sec/3_dataset}
\input{sec/4_method}

\input{sec/5_evaluation}

\input{sec/6_conclusion}
{
    \small
    \bibliographystyle{ieeenat_fullname}
    \bibliography{main}
}


\end{document}

%% file: sec/0_abstract.tex

\begin{abstract}

As the volume of video content online grows exponentially, the demand for moderation of unsafe videos has surpassed human capabilities, posing both operational and mental health challenges.
While recent studies demonstrated the merits of \modelnamefull\ (\modelname)  in various video understanding tasks, their application to multimodal content moderation, a domain that requires nuanced understanding of both visual and textual cues, remains relatively underexplored. In this work, we benchmark the capabilities of \modelname\ 
in brand safety classification, a critical subset of content moderation for safeguarding advertising integrity.
To this end, we introduce a novel, multimodal and multilingual dataset, meticulously labeled by professional reviewers in a multitude of risk categories. Through a detailed comparative analysis, we demonstrate the effectiveness of \modelname\ such as Gemini, GPT, and Llama in multimodal brand safety, and evaluate their accuracy and cost efficiency compared to professional human reviewers.  Furthermore, we present an in-depth discussion shedding light on limitations of \modelname\ and failure cases. We are releasing our dataset alongside this paper to facilitate future research on effective and responsible brand safety and content moderation.

\end{abstract}

%% file: sec/1_intro.tex
  \vspace{-2mm}
\section{Introduction}
Content moderation is a central pillar of trust and safety on social media platforms, playing a crucial role in protecting users, enforcing community standards, and preserving platform integrity. It has become an increasingly critical challenge, requiring significant manual effort and posing a substantial financial burden on organizations \cite{kuo-etal-2023-heroes, gillespie-etal-2023-scale}. The exponential growth of user-generated videos uploaded daily requires scalable, accurate, and cost-efficient moderation solutions. Moreover, reliance on human reviewers, while essential, is not without drawbacks. These individuals are often exposed to sensitive and disturbing content, leading to documented negative impacts on their well-being \cite{spence-etal-2023-psychological}. Recent advances in video understanding capabilities, particularly with \modelnamefull\ (\modelname) \cite{hu-etal-2024-fiova,he-etal-2024-annollm}, present a potential alternative or augmentation of traditional human moderation. The pursuit of scalable content moderation can mitigate the spread of harmful content and contribute to a safer online environment.

Our work focuses on the brand safety domain, an important subset of content moderation, particularly relevant in the advertising ecosystem. In this domain, advertisers specify general guidelines of content which they would prefer not to show up adjacent to, when their advertisements are displayed. Advertisers define content categories they wish to avoid; ranging from violent or adult-themed material to controversial political discourse. While general content moderation aims to identify and manage policy-violating content, brand safety is specifically concerned with aligning ad placements with advertiser preferences. Brand safety technology typically utilizes a hybrid approach of human review and machine learning to analyze multimodal signals, visual, audio, and text - and categorize content according to industry and brand-specific safety standards.

This presents the challenge of classifying unsafe and unsuitable video content at scale. Our main research questions are as follows:
\begin{itemize}
    \item \textbf{RQ1:} Can \modelnamefull\ (\modelname) perform accurately in brand safety tasks?
    \item \textbf{RQ2:}  
    How do the accuracy and cost-efficiency of \modelname\ compare to professional human reviewers?
\end{itemize}

Although \modelname\ have shown promise in various video understanding and annotation tasks \cite{hu-etal-2024-fiova,tan-etal-2024-large,he-etal-2024-annollm}, their application to multimodal content moderation, a domain that requires nuanced understanding of both visual and textual cues, remains relatively underexplored. Moreover, previous work focusing on Large Language Models (LLMs) for content moderation has focused mainly on textual data \cite{kumar-etal-2024-watch, kolla-etal-2024-llmmod}. However, such approaches fail to capture crucial visual cues present in video content, such as subtle gestures, the depiction of graphic scenes, or the context implied by the background environment. In contrast, in this work we study the application of \modelnamefull\ in a real-world setting and consider the multimodal aspects of video moderation including frames, thumbnail images, associated text, and automatically generated transcripts to better mimic the data seen by human moderators.

To address these research questions, we curated a novel dataset of thousands of labeled videos from the largest social media platforms. These videos were labeled by professional human reviewers, ensuring high quality and relevance. This dataset is unique in its multimodal composition, incorporating visual and textual data. The annotations by the reviewers specify the modality through which the risk signal manifests in the video, allowing us to study the explainability of the results and the contribution of each modality to the decision making. Another desirable property of the dataset is its multilingual nature, allowing us to evaluate the robustness of \modelname\ in content moderation across a dozen of languages.
A detailed description of the data set properties is provided in Section 3.  Its comprehensive nature, multimodal and multilingual aspects, and diverse risk categories allow for studies on domain adaptation and bias detection. We are releasing this dataset with the paper\footnotemark[1], hoping to spur future research on this crucial task.

In this work, we study both off-the-shelf and open-source \modelname\, including GPT~\cite{openai_gpt4o}, Gemini~\cite{deepmind_gemini}, and Llama~\cite{meta_llama}, across a diverse range of risk categories. We experiment with various techniques inspired by best practices in the literature, including chain-of-thought to improve the model’s reasoning ability. 
We evaluate the accuracy and cost-efficiency of these \modelname\ on our dataset and benchmark their performance against that of professional human reviewers. The results provide critical insights for the development of economically viable and highly effective content moderation strategies for social platforms.
Finally, we present a detailed discussion shedding light on limitations of MLLMs and failure cases, offering valuable insights for practitioners and researchers, and fostering a responsible approach to deploying AI in this domain. Overall, the key contributions of this work can be summarized as follows:
\begin{enumerate}
    \item We contribute a novel, multimodal and multilingual dataset, meticulously labeled by professional reviewers in a multitude of risk categories, which is made publicly available\footnote{
\url{https://mmcmd.zefr.com/}
} to spur future research on this important task. 
    \item We unveil the potential of \modelnamefull\ such as Gemini, GPT, and Llama in the task of multimodal brand safety, offering a path toward cost-efficient moderation systems that support safer online environments.
    \item We provide a detailed comparative analysis of the accuracy and cost efficiency of state-of-the-art \modelname, both off-the-shelf and open-source, compared to professional human reviewers.
\end{enumerate}


%% file: sec/2_related_work.tex
\section{Related Work}
Our work spans three closely related areas: (1) LLMs for annotation and video understanding, (2) Content moderation and brand safety methods, and (3) LLMs for content moderation. The following subsections present an overview of each of these areas.

\subsection{LLMs for annotation and video understanding}

The growing volume of user-generated video content requires efficient and scalable moderation techniques, driving research into the potential of AI-driven solutions. A critical component of this effort involves assessing how well \modelnamefull\ (\modelname) can understand and interpret video content in a manner comparable to human moderators. Hu et al. \cite{hu-etal-2024-fiova} address this challenge by introducing FIOVA, a novel benchmark designed to comprehensively evaluate the video understanding capabilities of \modelname\ through video captioning. 

A significant bottleneck in developing and deploying effective AI moderation systems is the availability of high-quality, labeled data. Tan et al. \cite{tan-etal-2024-large} provide a comprehensive survey of the use of large language models (LLMs) for data annotation and synthesis, highlighting their potential to automate and improve the annotation process. 
Similarly, He et al. \cite{he-etal-2024-annollm} propose AnnoLLM, an LLM-powered annotation system designed to potentially replace human annotators. AnnoLLM uses a novel "explain-then-annotate" approach, where LLMs first generate explanations for assigned labels and then use these explanations to annotate unlabeled data, achieving performance comparable to human annotators. 
While these studies demonstrate the potential of \modelname\ for image annotation,  our work extends these findings by conducting a comprehensive evaluation of the capabilities of \modelname\ in the domain of video moderation, which requires understanding of nuanced interactions between visual and textual elements.

\subsection{Content moderation and brand safety methods}

The growing scale of online content necessitates effective moderation and brand safety techniques. Kuo et al. \cite{kuo-etal-2023-heroes} highlight the challenges faced by volunteer moderators on platforms like Facebook, emphasizing the need for context-sensitive and personalized automated tools that maintain user agency. 
Gillespie et al. \cite{gillespie-etal-2023-scale} critically examine the push towards automated content moderation, questioning whether it is a necessary response to scale and raising concerns about over-reliance on technology and potential biases. The authors argue that relying solely on AI can lead to oversimplification and overlook crucial contextual nuances, potentially undermining the very values that content moderation aims to uphold. 

In the brand safety domain, Singhal et al. \cite{singhal-etal-2021-brand} explore using machine learning to redefine brand safety in programmatic advertising.  They study the techniques of Natural Language Processing (NLP), computer vision, and sentiment analysis to gauge content quality and context across various platforms. In addition, Korotkova et al. \cite{korotkova-etal-2023-beyond} show that common toxicity detection datasets are not enough for brand safety and demonstrate the need for building brand safety specific datasets.

In this work, we aim to address the limitations of relying solely on human moderators or unimodal AI systems by exploring the potential of \modelname\ to understand and classify video content in a manner that mirrors the complex decision-making processes of human reviewers. Moreover, following up on prior work, we present a detailed discussion shedding light on limitations of \modelname\ and failure cases, offering insights into the design of more effective hybrid human-AI moderation systems.

\subsection{LLMs for content moderation}
Building on the increasing adoption of AI methods in content moderation and recent advances in Large Language Models, prior research has explored the use of LLMs for content moderation, primarily focusing on text-based data. Kumar et al. \cite{kumar-etal-2024-watch} evaluate various LLMs for rule-based community moderation and toxicity detection, observing that LLMs outperform traditional toxicity classifiers. Kolla et al. \cite{kolla-etal-2024-llmmod} investigate LLMs for identifying rule violations on Reddit, finding limitations in their ability to handle complex reasoning. Huang et al. \cite{huang-etal-2024-legit} propose a shift from purely accuracy-based evaluation towards using LLMs to provide justifications and to facilitate user participation.

These studies provide valuable insights into the capabilities and limitations of LLMs in content moderation. However, they largely neglect the crucial visual aspects of video content. Our work directly addresses this gap by investigating the application of multimodal LLMs for video content moderation. To the best of our knowledge, this is the first study to systematically evaluate the effectiveness of Multimodal Large Language Models in brand safety classification tasks. We present a novel multimodal dataset and leverage both visual and textual information derived from videos, including frames, thumbnails, associated text, and transcripts, to assess the potential of \modelname\ in detecting unsafe and unsuitable content. 

\label{sec:formatting}

%% file: sec/3_dataset.tex
\section{Dataset}
The dataset we present in this paper directly addresses a critical gap in multimodal content moderation research: a high-quality, publicly available benchmark dataset specifically designed for classifying multimodal content. While existing public datasets are typically modality-specific, such as textual datasets for hate speech and toxicity detection \cite{davidson-etal-2017-hate, korotkova-etal-2023-beyond}, or visual datasets for identifying adult content \cite{duy-etal-2022-adult}, our dataset captures the interplay between textual and visual cues that characterize real-world content moderation challenges. In contrast to recent works that study the accuracy of \modelname\ on text and images separately \cite{kumar-etal-2024-watch}, our work benefits from the use of the combined multimodal nature of video data. Widely used public datasets for video classification, such as the YouTube-8M benchmark \cite{haija-etal-2016-youtube8m}, are typically curated to exclude sensitive content, making them unsuitable as benchmarks for evaluating content moderation systems. In addition, as demonstrated by recent work \cite{korotkova-etal-2023-beyond}, it is necessary to build brand safety specific datasets.

The primary objective of this dataset is to provide a resource for researchers to develop and evaluate \modelname\ for effective and responsible brand safety and content moderation. It provides a benchmark for the comprehensive evaluation of AI behavior in realistic, high-stakes environments, including assessments of modality attribution and alignment with human judgment. We acknowledge that releasing a dataset with potentially sensitive content requires responsible usage. We have taken protective measures to exclude extremely graphic or explicit content and we shall include ethical guidelines alongside the release. 

\subsection{Data collection}


The dataset we present in this work consists of 1500 videos, evenly distributed across three risk categories, with 500 videos randomly sampled from each category between July 1st, 2024, and March 1st, 2025:
\begin{enumerate}
    \item Drugs, Alcohol and Tobacco (DAT)
    \item Death, Injury and Military Conflict (DIMC)
    \item Kid's Content
\end{enumerate}
These categories are standard risk categories in content moderation. They were selected to offer a diverse mix of textual and visual cues. Moreover, they consist of more granular sub-categories (e.g. Drugs, Alcohol and Tobacco) - allowing us to observe the robustness of the \modelname\ classification for diverse content.
Kid's Content poses a risk for certain advertisers, such as alcohol brands or pharmaceutical companies, that seek to avoid targeting children and, in some cases, are legally prohibited from doing so.

The videos were labeled by professional human reviewers using a labeling platform. Once the reviews are completed, a senior team member verified a significant sample of the reviews to measure the accuracy of the reviewers and to ensure high quality.
For each video, the review process considers multiple modalities: text, audio, and visuals, including the thumbnail and video frames. Videos on social media span minutes to hours. In this work, the reviewers review the whole duration of the content, or the first 60 seconds in case the video is longer than 60 seconds. Our work can be extended by applying similar techniques for long-format videos. 

\subsection{Data characteristics}

Our dataset consists of 1500 videos, evenly distributed across three risk categories, with 500 videos per category, as shown in  Table~\ref{tab:data_labels}. Overall, the dataset exhibits a relatively balanced distribution between positive (risky) and negative (safe) videos. Specifically, we have a total of 823 positive videos and 677 negative videos. Notably, the Kids category has the highest proportion of risky videos, whereas DAT and DIMC have a more even split. 

\begin{table}[h!]
\centering
\begin{tabular}{|c|c|c|c|c|}
\hline
\textbf{Label} & \textbf{DAT} & \textbf{DIMC} & \textbf{Kids} & \textbf{Total} \\
\hline
Positive & 241 & 276 & 306 & 823 \\
Negative & 259 & 224 & 194 & 677 \\
\hline
Total & 500 & 500 & 500 & 1500 \\
\hline
\end{tabular}
\caption{Video Count by Label and Category}
  \vspace{-2mm}
\label{tab:data_labels}
\end{table}

\begin{figure}[ht]
    \centering
    \includegraphics[width=\columnwidth]{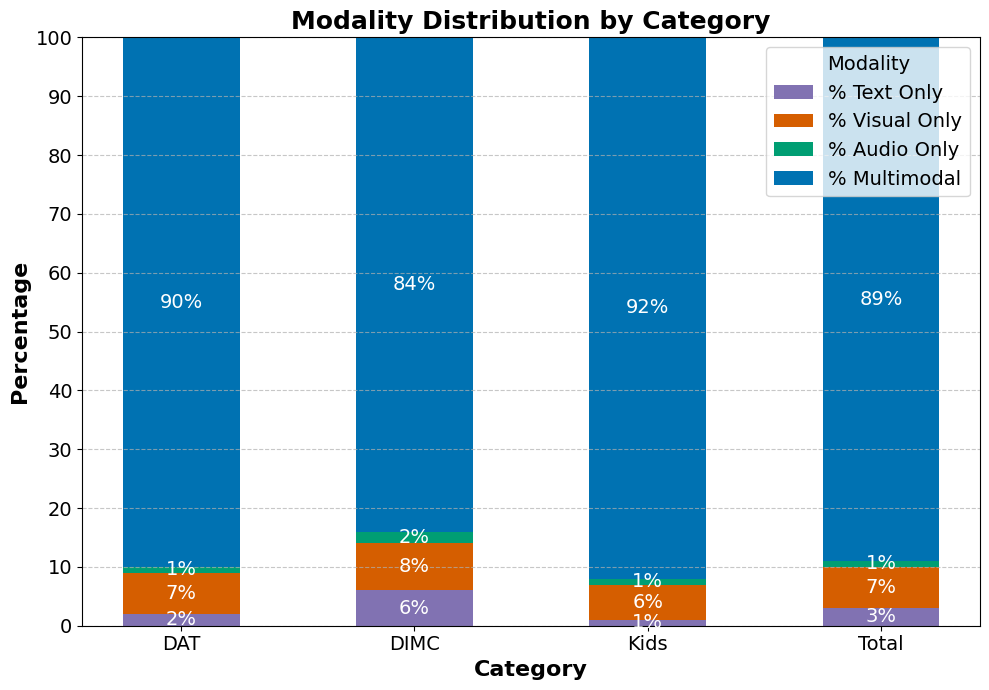}
    \caption{Percentage Distribution of the Risk Signals Present in Videos by Modality for each Category.}
    \label{fig:stacked_bar_chart}
\end{figure}

In addition to the labels, the dataset includes detailed annotations by the professional reviewers. These annotations specify, with regard to the risky videos, the modality of the risk signals that are present in the videos: text, visual, audio (transcript), or any combination thereof (multimodal). In Figure \ref{fig:stacked_bar_chart}, we present the percentage distribution of the risk signals present in the videos by modality (Text only, Visual only, Transcript only, and Multimodal) for each category. For the vast majority of videos, the risk signals are present in multiple modalities (89\% in total), highlighting the multimodal nature of the content moderation task. The distribution of risk by modality varies slightly across the categories, with the Kids category having the largest share of Multimodal risk (92\%),  while the DIMC category has the highest proportion of Visual Only (8\%) and Text Only (6\%) risk. Notably, for a significant percentage of videos, the risk signals are present only in the visuals or only in the transcript, demonstrating the opportunity to achieve higher accuracy by using \modelname\ rather than text-only LLMs.

\begin{figure}[ht]
    \centering
    \includegraphics[width=\columnwidth]{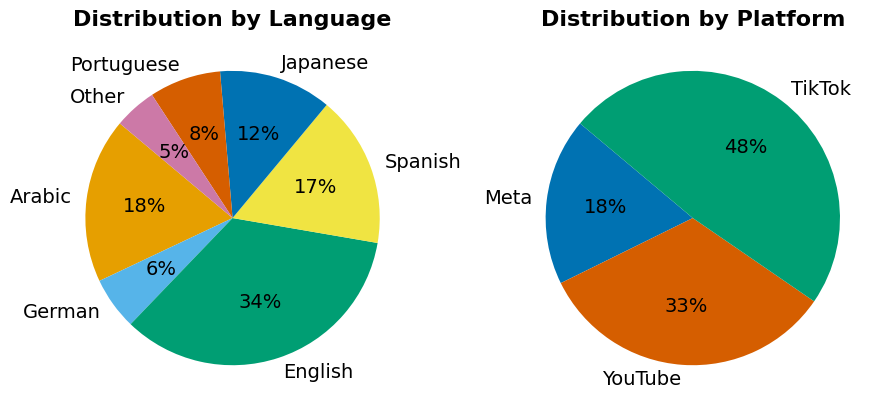}
    \caption{Percentage Distribution by Language and by Platform.}
    \vspace{-2mm}
    \label{fig:pie_videos}
\end{figure}

A key strength of this dataset is its multi-platform and multilingual nature, reflecting the diverse landscape of social media content.  Figure \ref{fig:pie_videos} presents the distribution of videos by platform and language.
The dataset encompasses four major social media platforms: TikTok (48.5\%), YouTube (33.1\%), and Meta (Facebook and Instagram combined, 18.4\%). 
Furthermore, it contains videos in a dozen languages, with English comprising 34.5\% of the content. Other prominent languages include: Arabic (18.1\%), Spanish (16.7\%), Japanese (12.4\%), Portuguese (7.9\%) and German (5.8\%). This dataset offers an opportunity to evaluate the reliability and reasoning alignment of AI systems across linguistic and cultural boundaries.

%% file: sec/4_method.tex
\section{Method}

Our method is designed to evaluate the performance of \modelnamefull\ (\modelname) in detecting unsafe video content across various risk categories and multiple modalities, addressing research questions \textbf{RQ1} and \textbf{RQ2}. The process consists of two primary stages: (1) Data Preparation, where we extract and process multimodal data from video content; and (2) Zero-Shot Classification, where we assess the inherent capabilities of \modelname\ in brand safety moderation tasks using prompting techniques inspired by best practices in the literature.

We study both off-the-shelf and open-source \modelname\, including models from the GPT~\cite{openai_gpt4o}, Gemini~\cite{deepmind_gemini}, and Llama~\cite{meta_llama} families. These models were selected for their advanced multimodal capabilities, representing state-of-the-art performance in vision-language understanding. \texttt{GPT-4o}~\cite{openai_gpt4o}, being a widely recognized and versatile language model, excels in handling text-based queries and is capable of working with visual content when combined with image inputs. \texttt{Gemini-2.0-Flash}~\cite{deepmind_gemini}, on the other hand, supports both textual and visual inputs natively, and is particularly optimized for high-performance multimodal tasks. 
\texttt{Llama-3.2-11B-Vision}~\cite{meta_llama} is an open-source model specifically designed for tasks involving both language and vision, offering transparency and flexibility for fine-tuning.
Additionally, we analyze compact variants of the Gemini and GPT models, specifically \texttt{GPT-4o-mini} and \texttt{Gemini-2.0-Flash-Lite}, to provide a comprehensive accuracy-cost tradeoff analysis. Furthermore, we evaluate the previous-generation \texttt{Gemini-1.5-Flash} model to assess cross-generational differences.

\subsection{Data preparation}
The inputs to our method include textual, auditory, and visual signals. For each video in our dataset, we use the YouTube, Meta and TikToK APIs to collect the video title, description, if available, a thumbnail image, and the MP4 video file. The video is trimmed to the first 60 seconds, in alignment with the review process described in Section 3, and individual frames are extracted. Initial experiments with a one-frame-per-second sampling rate resulted in significant frame redundancy. To improve cost-efficiency and reduce computational overhead, we used the open-source library PySceneDetect\cite{PySceneDetect} which detects shot changes in videos to extract key frames that represent distinct visual content.  In addition to visual data, audio transcripts are extracted using OpenAI's Whisper\cite{WhisperOpenAI}. Specifically, we utilize the whisper-large-v3-turbo model, an optimized version of large-v3 that offers faster transcription speed with a minimal degradation in accuracy.

The text, thumbnail images, and videos are supplied to the inputs of each of the \modelname\, alongside the prompt instructions. The combination of key frames, audio transcripts, and text allows the \modelname\ to more accurately classify the content in the videos.

\subsection{Zero-shot classification}
Following the data preparation, to evaluate the inherent capabilities of \modelname\ without task-specific training, we perform zero-shot classification using carefully designed prompts.  The prompt design aims to optimize performance and ensure consistent content moderation based on both visual and textual content. We treat zero-shot classification not just as a performance baseline but as an evaluation of the reasoning capabilities of MLLMs acting autonomously. By providing only prompts and inputs without training, we can assess the robustness, interpretability, and decision consistency of these models.

To achieve this, we implemented prompting techniques inspired by best practices in the literature.
First, we utilized instruction-based prompting to provide clear, explicit guidance on how the model should process the content, helping it consistently follow the task-specific steps \cite{brown-etal-2020-learners}. The prompt instructs the model to consider both visual and auditory elements of the video, as well as the textual content, reinforcing a multimodal approach to decision making. To maintain consistency and interpretability, we applied format-enforced prompting, constraining responses to a strict "YES" or "NO" format \cite{schick-etal-2020-exploiting}. Additionally, we incorporated a persona-based approach, instructing the model to act as an expert brand safety moderator, which helps anchor the model in the context of the task and guide its decision-making process \cite{brown-etal-2020-learners}. We embedded elements of chain-of-thought (CoT) by instructing the model to break down the text step-by-step before making a final decision, which is shown to improve reasoning accuracy in complex tasks \cite{wei-etal-2022-cot}. The prompt and additional reproducibility details are available in the GitHub page\footnotemark[1].
These techniques collectively aim to optimize the model's understanding and handling of the multimodal content while supporting clarity and consistency in its final output. While these techniques align with established methodologies in the literature, further empirical evaluation is needed to assess their effectiveness in the context of multimodal content moderation.

For each risk category, we defined a set of rules formalized as binary questions (yes/no) according to the policy. A single binary question is sent for each video, per risk category, and the model provides an assessment. For example, "does the video feature alcohol?", "does the video feature nudity?", and more. Finally, we applied a boolean logic on top of all questions to create the final prediction for each video. For example, if any of the questions regarding Drugs, Alcohol or Tobacco is answered as "YES", then the model decision is positive for that category.

We use the APIs of Google's Gemini and OpenAI's GPT models with batch requests.
In addition, we evaluate the Llama-Vision model on a local machine with A100 GPUs.
We configure the safety settings, if available, to BLOCK\_NONE - to avoid rejections by the API due to unsafe content.
When calling the models, we provide text, thumbnail images, key frames extracted using PySceneDetect, and transcripts extracted using Whisper as part of the prompt request. Notably, for Gemini, we provide the MP4 video directly, leveraging its native video processing capabilities, while for GPT and Llama, which handle only image inputs, we provide the extracted key frames. 




%% file: sec/5_evaluation.tex
\begin{table*}[h]
    \centering
    \resizebox{\textwidth}{!}{ 
    \begin{tabular}{lccc|ccc|ccc|ccc} 
        \toprule
        & \multicolumn{3}{c|}{Drugs, Alcohol and Tobacco} & \multicolumn{3}{c|}{Death, Injury and Military Conflict} & \multicolumn{3}{c|}{Kids} & \multicolumn{3}{c}{Overall} \\
        Model & Precision & Recall & F1-score & Precision & Recall & F1-score & Precision & Recall & F1-score & Precision & Recall & F1-score \\
        \midrule
        GPT-4o                & 0.94 & 0.92 & 0.93 & 0.94 & 0.64 & 0.76 & 0.93 & 0.92 & 0.93 & 0.94 & 0.83 & 0.87 \\
        GPT-4o-mini           & 0.94 & 0.93 & 0.93 & 0.91 & 0.72 & 0.80 & 0.92 & 0.90 & 0.91 & 0.92 & 0.85 & 0.88 \\
        Gemini-1.5-Flash      & 0.86 & 0.98 & 0.92 & 0.86 & 0.92 & 0.89 & 0.84 & 0.98 & 0.90 & 0.86 & 0.96 & 0.90 \\
        Gemini-2.0-Flash      & 0.83 & 0.99 & 0.91 & 0.86 & 0.95 & 0.91 & 0.83 & 0.99 & 0.90 & 0.84 & 0.98 & 0.91 \\
        Gemini-2.0-Flash-Lite & 0.86 & 0.98 & 0.91 & 0.88 & 0.91 & 0.90 & 0.87 & 0.97 & 0.92 & 0.87 & 0.95 & 0.91 \\
        Llama-3.2-11B-Vision  & 0.88     & 0.87     & 0.87     & 0.89     & 0.76     & 0.82     & 0.85     & 0.94     & 0.89     & 0.87 & 0.86 & 0.86 \\
        Human                 & 0.98 & 0.98 & 0.98 & 0.98 & 0.96 & 0.97 & 0.99 & 0.97 & 0.98 & 0.98 & 0.97 & 0.98 \\
        \bottomrule
    \end{tabular}
    } 
    \caption{Precision, Recall, and F1-score per Model and Category, vs. Human Reviewers.}
        \vspace{-2mm}
    \label{tab:metrics}
\end{table*}

\section{Evaluation}
In this section, we provide a comprehensive evaluation of the accuracy and cost-effectiveness of state-of-the-art  (\modelname) in comparison to professional  reviewers. We present a detailed analysis of the model accuracy per each risk category, alongside a breakdown of costs per model and modality, and explore the trade-off between accuracy and cost. Additionally, we examine the limitations and failure cases of \modelname\, offering insights into their practical applications and highlighting areas for improvement.

\subsection{Accuracy}


The results presented in Table~\ref{tab:metrics} highlight the comparative performance of state-of-the-art \modelnamefull\ (\modelname) alongside the performance of human reviewers across three categories - Drugs, Alcohol and Tobacco (DAT); Death, Injury and Military Conflict (DIMC); and Kids Content - as well as an overall assessment. 
Human performance serves as the benchmark, measured by a senior team member who verified a representative sample of the reviews. 
The metrics used for evaluation are precision, recall, and F1-score.

Among the \modelname\, the Gemini models emerge as the best overall models, outperforming the others in terms of F1-score. Specifically, Gemini-2.0-Flash and Gemini-2.0-Flash-Lite achieve the highest overall F1-score of 0.91, while Gemini-1.5-Flash follows closely behind with F1-score of 0.90. The GPT-4o and GPT-4o-mini models also perform strongly with F1 scores of 0.87 and 0.88, respectively. Lastly, Llama-3.2-11B-Vision lags behind, attaining the lowest F1-score (0.86). The performance of the models varies across categories. The GPT models achieve the highest F1-scores in DAT and Kids, however, they struggle significantly with DIMC, where recall drops notably (e.g., GPT-4o achieves only 0.64 recall). In contrast, the Gemini models offer a more balanced performance across all categories, with recall scores consistently exceeding 0.90.
Llama-3.2-11B-Vision generally underperforms except for the DIMC category where it outpeforms the GPT models.

Interestingly, compact versions of models do not exhibit significant performance degradation compared to their larger counterparts. Gemini-2.0-Flash-Lite performs on par with Gemini-2.0-Flash, while GPT-4o-mini marginally outperforms GPT-4o overall. Additionally, while Gemini-2.0-Flash exceeds the performance of Gemini-1.5-Flash, the gap is relatively small. This suggests that model size alone is not the primary determinant of classification accuracy in multimodal brand safety and content moderation. These findings align with recent advancements in model distillation, where smaller models are optimized to retain the performance of larger architectures while being computationally efficient. Furthermore, while larger models generally have a broader world knowledge base, this does not always translate directly to better task-specific performance. Content moderation tasks may rely more on pattern recognition and classification capabilities rather than extensive general knowledge. 

An interesting insight from the table is the trade-off between precision and recall across different model architectures. The Gemini models prioritize recall, capturing a broader spectrum of relevant content, whereas GPT-4o and its mini variant maintain higher precision, reducing false positives. Combining these approaches can optimize performance based on the relative costs of false positives vs. false negatives.

When compared to human reviewers, all \modelname\ exhibit a noticeable performance gap, with humans consistently achieving higher F1-scores across all categories. Humans outperform the models by a significant margin in most cases, achieving an overall F1-score of 0.98. These results underscore the effectiveness of \modelname\ in automating content moderation but also highlight the continued superiority of human reviewers in accuracy, particularly in more complex or nuanced classifications where context and deep understanding are required. 

\begin{table}[h]
    \centering
    \footnotesize
    \begin{tabular}{|l|c|c|c|c|c|}
        \hline
        \textbf{Model} & \textbf{Modality} & \textbf{DAT} & \textbf{DIMC} & \textbf{Kids} & \textbf{All} \\
        \hline
        GPT-4o   & Text       & 0.91 & 0.77 & 0.92 & 0.87 \\
                        & Multimodal      & 0.93 & 0.76 & 0.93 & 0.87 \\
        \hline
        GPT-4o-mini & Text       & 0.89 & 0.78 & 0.90 & 0.86 \\
                        & Multimodal      & 0.93 & 0.80 & 0.91 & 0.88 \\
        \hline
        Gemini-1.5-Flash & Text      & 0.89 & 0.75 & 0.89 & 0.84 \\
                         & Multimodal      & 0.92 & 0.89 & 0.90 & 0.90 \\
        \hline
        Gemini-2.0-Flash & Text      & 0.90 & 0.79 & 0.90 & 0.86 \\
                        & Multimodal      & 0.91 & 0.91 & 0.90 & 0.91 \\
        \hline
        Gemini-2.0-Flash-Lite & Text  & 0.87 & 0.81 & 0.90 & 0.86 \\
                        & Multimodal      & 0.91 & 0.90 & 0.92 & 0.91 \\
        \hline
        Llama-3.2-11B-Vision & Text & 0.85 & 0.81 & 0.87 & 0.84 \\
                        & Multimodal & 0.87 & 0.82 & 0.89 & 0.86 \\
        \hline
    \end{tabular}
    \caption{F1-Scores of models: Text vs. Multimodal per Category.}
    \label{tab:model_performance}
\end{table}

\begin{table*}[h]
    \centering
    \footnotesize
    \setlength{\tabcolsep}{3pt}
    \renewcommand{\arraystretch}{1.0}
    \begin{tabular}{lrrrrrr}
        \hline
        \textbf{Cost Factor} & \textbf{GPT-4o-mini} & \textbf{GPT-4o} & \textbf{Gemini-1.5-Flash} & \textbf{Gemini-2.0-Flash} & \textbf{Gemini-2.0-Flash-Lite} & \textbf{Llama-3.2-11B-Vision} \\
        \hline
        text\_tokens & 22,352,010 & 22,351,674 & 24,079,694 & 24,080,112 & 27,240,271 & 123,866,395 \\
        text\_chars & 0 & 0 & 84,278,928 & 84,280,339 & 24,240,103 & 0 \\
        \hline
        transcript\_tokens & 7,632,550 & 7,632,526 & 7,372,181 & 7,372,202 & 7,372,306 & 197,574,338 \\
        transcript\_chars & 0 & 0 & 22,162,601 & 22,162,656 & 22,160,690 & 0 \\
        \hline
        image\_tokens & 304,949,505 & 304,947,130 & 5,627,496 & 37,927,806 & 5,247,624 & 7,786,464,000 \\
        video\_tokens & 0 & 0 & 756,985,175 & 690,622,140 & 792,261,016 & 0 \\
        \hline
        \textbf{total\_input\_tokens} & \textbf{334,934,065} & \textbf{334,931,330} & \textbf{794,064,546} & \textbf{760,002,260} & \textbf{832,121,217} & \textbf{8,107,904,733} \\
        \hline
        output\_tokens & 124,476 & 124,474 & 124,471 & 62,284 & 63,310 & 647,176,218 \\
        output\_chars & 0 & 0 & 196,459 & 134,681 & 134,214 & 0 \\
        \hline
        cost\_text & \$1.71 & \$28.56 & \$0.80 & \$1.59 & \$0.23 & \$6.88 \\
        cost\_transcript & \$0.25 & \$9.54 & \$0.03 & \$0.07 & \$0.15 & \$11.17 \\
        cost\_image & \$22.87 & \$381.18 & \$0.22 & \$2.11 & \$1.05 & \$440.60 \\
        cost\_video & \$0.00 & \$0.00 & \$27.13 & \$52.50 & \$26.27 & \$0.00 \\
        \hline
        total\_cost & \$24.83 & \$419.29 & \$28.18 & \$56.27 & \$27.70 & \$458.65 \\
        \hline
        Cost per million input tokens & \$0.074 & \$1.252 & \$0.035 & \$0.074 & \$0.033 & \$0.057 \\
        \hline
    \end{tabular}
    \caption{Model Token Usage and Cost Breakdown per Modality.}
        \vspace{-3mm}
    \label{tab:cost_breakdown}
\end{table*}

Notably, the recall scores of the Gemini models are on par and even exceed the recall of human reviewers in the DAT and Kids categories. Additionally, Gemini-2.0-Flash achieves a recall of 0.95 in the DIMC category, only slightly trailing human reviewers (0.96). These results suggest the potential efficacy of a hybrid human-AI review system, where AI serves as an initial filter to expedite content moderation, followed by human oversight to ensure precision and minimize classification errors.


The results in Table~\ref{tab:model_performance} reveal a consistent trend: multimodal models outperform their text-only counterparts across all  categories and model versions, with a single exception in the DIMC category, where the multimodal GPT-4o slightly underperforms compared to its text-only variant.

A notable observation is the magnitude of improvement from text-only to multimodal processing. The Gemini models exhibit a significantly larger performance gain in the multimodal setting compared to the GPT models. This disparity suggests two potential explanations: (1) the GPT models possess stronger textual reasoning capabilities, reducing their dependence on additional modalities, or (2) the Gemini models exhibit superior visual understanding, likely due to their native ability to process video content, whereas GPT models are limited to individual image frames. The latter hypothesis aligns with Gemini’s design, which explicitly supports richer temporal and spatial comprehension in multimodal tasks.
Overall, the results highlight that while text-only models remain competitive, multimodal models offer a significant edge in handling more complex or context-dependent tasks, underlining the importance of integrating multimodal inputs (e.g., text, visual, and audio data) to achieve higher classification performance.

\begin{table}[h]
    \centering
    \begin{tabular}{lcc} 
        \toprule
        Model                    & F1-Score  & Total Cost  \\  
        \midrule
        GPT-4o                   & 0.87      & \$419   \\  
        GPT-4o-mini              & 0.88      & \$25   \\  
        Gemini-1.5-Flash         & 0.90      & \$28     \\  
        Gemini-2.0-Flash         & 0.91      & \$56    \\  
        Gemini-2.0-Flash-Lite    & 0.91      & \$28    \\  
        Llama-3.2-11B-Vision     & 0.86      & \$459  \\  
        Human                    & 0.98         & \$974   \\  
        \bottomrule
    \end{tabular}
    \caption{F1-Scores and Total Cost per Model.}
    \vspace{-2mm}
    \label{tab:cost_efficiency}
\end{table}

\begin{table*}[h!]
\centering
\renewcommand{\arraystretch}{1} 
\footnotesize 
\resizebox{\textwidth}{!}{%
\begin{tabular}{|c|c|c|c|c|c|c|c|c|}
\hline
\textbf{Example URL} & \textbf{4o} & \textbf{4o-mini} & \textbf{1.5-Flash} & \textbf{2.0-Flash} & \textbf{2.0-Flash-Lite} & \textbf{Llama-3.2} & \textbf{Human} & \textbf{Final Label} \\
\hline
\href{https://www.tiktok.com/@d/video/7292396554105687303}{tiktok.com/@d/video/7292396554105687303} & DAT & DAT & DAT & DAT & DAT & DAT & No DAT & No DAT \\
\href{https://www.tiktok.com/@d/video/7342745883508821290}{tiktok.com/@d/video/7342745883508821290} & No DAT & No DAT & No DAT & DAT & No DAT & DAT & No DAT & No DAT \\
\href{https://www.youtube.com/watch?v=wGubE4MeSU8}{youtube.com/watch?v=wGubE4MeSU8} & No Kids & No Kids & Kids & Kids & Kids & No Kids & No Kids & No Kids \\
\href{https://www.youtube.com/watch?v=wTAnk8SVO6U}{youtube.com/watch?v=wTAnk8SVO6U} & No Kids & No Kids & No Kids & No Kids & No Kids & No Kids & Kids & Kids \\
\href{https://www.tiktok.com/@sskelefilms/video/7432660344310402337}{tiktok.com/@d/video/7432660344310402337} & DIMC & DIMC & DIMC & DIMC & DIMC & DIMC & No DIMC & DIMC \\
\hline
\end{tabular}
}
\caption{Example Videos: Model Classifications Compared to Human Labels.}
\vspace{-3mm}
\label{tab:examples}
\end{table*}


\subsection{Cost-Efficiency}

Table~\ref{tab:cost_breakdown} presents a detailed cost breakdown of each model across different modalities, including text, visual, and transcript inputs, as well as the overall cost. The breakdown is derived from the token consumption of each model, aggregated across the three risk categories, with costs computed using API pricing for OpenAI's GPT \cite{gpt_pricing} and Google's Gemini \cite{gemini_pricing}. 
For text and transcript inputs, the pricing model differs between GPT and Gemini. GPT models are charged based on the number of tokens, whereas Gemini models use character-based pricing. For visual inputs, GPT processes images extracted from key frames, which are reflected in the reported image\_tokens. In contrast, Gemini processes video files directly in MP4 format, leading to the reported video\_tokens.

To compute the costs, we multiply the number of tokens (or characters, where applicable) by the corresponding API rates for each modality. 
To calculate the costs of the Llama-3.2-11B-Vision model, we measure the total inference duration of the model and multiply it by the compute cost of a machine equipped with a CPU and an A100 GPU, based on Google’s Vertex AI pricing \cite{vertex_pricing}. The cost attribution per modality is determined according to the distribution of input tokens. Finally, the total cost for each model is then obtained by summing costs across all modalities.

Notably, the token consumption of Llama-3.2-11B-Vision is several orders of magnitude higher than that of other models. This discrepancy is likely due to differences in model architecture and serving infrastructure — Gemini, GPT, and other large models benefit from optimized serving systems that incorporate caching and batching techniques. However, when analyzing cost per million input tokens, Llama-3.2-11B-Vision remains comparable to the more compact variants of GPT and Gemini.

It can be observed that the visual component accounts for the majority of the cost across all models. This highlights the trade-off between achieving superior classification performance through multimodal inputs, as presented in Section 5.1, and significantly higher costs compared to text-only LLMs. Practitioners should carefully consider the optimal balance between accuracy and cost, depending on their preferences.


Table~\ref{tab:cost_efficiency} summarizes the F1-scores and total costs of all models compared to human reviewers. The cost of human reviewers was computed by multiplying the total review time by their hourly rate (USD). Among AI models, Gemini-2.0-Flash and Gemini-2.0-Flash-Lite achieve the highest F1-score (0.91) at relatively low costs (\$56 and \$28, respectively). The cheapest model, GPT-4o-mini (\$25), performs slightly worse (F1=0.88) but remains highly cost-effective. Conversely, Llama-3.2-11B-Vision and GPT-4o are the least efficient, incurring significantly higher costs (\$459 and \$419) while achieving lower F1-scores (0.86 and 0.87). 
When compared to human cost, all models provide a significantly cheaper alternative, highlighting the potential for cost reduction using hybrid human-AI systems.
Furthermore, we observe that the compact models, GPT-4o-mini and Gemini-2.0-Flash-Lite, provide a significantly cheaper alternative compared to their larger counterparts, with no degradation in accuracy, presenting a very attractive path for cost-efficient deployment.






\subsection{Limitations and failures}

Our evaluation revealed several failure cases in current \modelname\, including incorrect associations, insufficient contextual understanding, and language support challenges.  Examples of misclassifications are presented in Table~\ref{tab:examples}. A primary limitation observed was the models' tendency to make incorrect associations, leading to misclassifications. For example, a video discussing addiction to caffeine in Japanese was incorrectly classified by all models as DAT (Drugs, Alcohol, and Tobacco), due to the flawed association of the term addiction, and potential challenges of contextual understanding in Japanese compared to English. Generally, the models exhibited poorer performance on non-English content, raising concerns about language bias and equitable treatment of global content. 

Furthermore, the models exhibited mixed results when presented with ambiguous terms. A video titled "grass dealer" (referring to a lawn care company) was incorrectly classified by Gemini-2.0-Flash and Llama as DAT. Similarly, the Gemini models incorrectly labeled adult content about “sex toys” as kids content, likely due to the word “toys”. In another case, all models incorrectly classified a video of a person in a furry costume telling a children’s story as No Kids, suggesting a lack of nuanced understanding of video content. Interestingly, there were a few cases where human reviewers made incorrect judgments while the models produced accurate results. In one example featuring a fictional serial killer character (Dexter), all models correctly classified it as Death, Injury, and Military Conflict. The reviewer initially classified it as No DIMC and it was corrected by the senior reviewer. This highlights the potential for models to incorporate world knowledge effectively.

Our study was designed to establish a foundational performance baseline for \modelname\ in brand safety, and this focus defines the limitations of our work. First, our dataset includes three core risk categories, though further research is needed to assess generalizability to a wider array of content domains. Second, our analysis focuses on the first 60 seconds of video, which ensures a consistent evaluation setup, but may miss important context present in long-form content. Finally, our evaluation focuses on zero-shot performance to assess the models’ out-of-the-box capabilities, which provides a reproducible baseline, but does not capture the full potential under fine-tuning or domain-specific training. These limitations outline clear and actionable directions for future work.

%% file: sec/6_conclusion.tex
\section{Conclusion}

In this work, we introduced a novel dataset that fills a critical gap in multimodal brand safety and content moderation research: a high-quality, publicly available benchmark featuring multimodal and multilingual videos across multiple platforms and risk categories. 
We studied off-the-shelf and open-source MLLMs: GPT, Gemini, and Llama 
- and demonstrated their effectiveness in scaling content moderation with promising cost-reduction potential in hybrid human-AI systems.
We highlighted the effectiveness of combining textual, visual, and auditory signals - with multimodal models consistently outperforming text-only methods. We showed that the compact MLLMs offer a significantly cheaper alternative compared to their larger counterparts without sacrificing accuracy.
However, human reviewers remain superior in accuracy, particularly in complex or nuanced classifications. Our analysis further revealed 
key limitations of MLLMs including incorrect associations and challenges in language support, underscoring areas for future improvement.
For future work, we aim to explore fine-tuning \modelname\ to enhance performance in brand safety tasks, and evaluating the cost-efficiency of \modelname\ for long-format videos. We have made the dataset publicly available and plan to expand it with additional risk categories to spur future research on this crucial task.